%% file: Main.tex
\documentclass{article}
\usepackage{iclr2027_conference,times}

\input{math_commands.tex}

\usepackage{amsmath,amssymb}
\usepackage{booktabs}
\usepackage{graphicx}
\usepackage{url}
\usepackage{comment}
\usepackage{multirow}
\usepackage{xcolor}
\usepackage{tabularx}
\usepackage{algorithm}
\usepackage{algpseudocode}
\usepackage{placeins}
\usepackage{hyperref}
\usepackage{comment}
\definecolor{familyshade}{RGB}{250,246,247}
\usepackage[table]{xcolor}
\title{FamilySpace: From Emergent Chemical Semantics to Controllable Crystal Generation}

\author{
    Haowei Hua$^{1}$, Irwin King$^{2}$, Wanyu Lin$^{1}$\thanks{ Correspondence to: \texttt{wanyu.lin@polyu.edu.hk}}\\
    $^{1}$Department of Computing, The Hong Kong Polytechnic University \\ 
    $^{2}$Department of Computer Science and Engineering, The Chinese University of Hong Kong
}
\iclrfinalcopy  

\begin{document}

\maketitle

\begin{abstract}
Latent crystal generative models typically combine a variational autoencoder (VAE) for learning atom-wise representations with a flow model that transports a simple source distribution to the learned latent distribution. However, previous studies have mainly focused on the quality of generated crystals, leaving the structure and generative utility of the atom-wise latent space largely unexplored. 
In this work, we show that even a vanilla crystal VAE implicitly organizes atom-wise latent representations according to elemental families, with different families occupying largely separated regions of the latent space. We term this phenomenon {\em emergent chemical semantics}. This structure suggests that sampling from a family-specific latent region can bias the decoder toward atoms from the corresponding elemental family. Building on this insight, we propose FamilySpace, a framework for controlling the elemental-family composition of generated crystals. FamilySpace assigns each elemental family a distinct source component, forming a family-indexed multi-source distribution, and trains a flow model using source-target pairs matched by family. In inference, a user-specified family template selects the source component for each atom-wise representation, enabling direct control over the family composition of generated crystals. Experiments on MP-20 and MPTS-52 show that FamilySpace achieves competitive crystal-generation performance while matching prescribed elemental-family compositions. Moreover, selected family templates increase the fraction of generated crystals predicted to be stable or metastable, demonstrating that latent-space composition control can support targeted exploration of promising chemical regions without additional optimization techniques such as reinforcement learning. Code is available at \url{https://github.com/hhw11/FamilySpace}

\end{abstract}

\input{Introduction}

\input{Related_work}
\input{Preliminaries}

\input{Methodology}

\input{Experiments}

\section{Conclusion}
\label{sec:conclusion}

In this work, we investigated the geometric organization of atom-wise latent representations learned by a vanilla crystal VAE and showed that they exhibit emergent chemical semantics: representations from the same elemental family form coherent neighborhoods, while different families occupy largely separated regions of the latent space. Building on this observation, we introduced FamilySpace, which converts this latent organization into a family-indexed multi-source distribution and uses family-matched source--target coupling to enable controllable latent flow generation. Experiments on MP-20 and MPTS-52 demonstrate that FamilySpace maintains competitive crystal-generation performance while accurately following prescribed elemental-family compositions. Moreover, chemically motivated family templates can direct generation toward regions with higher fractions of candidates predicted to be stable or metastable, without additional optimization such as reinforcement learning. These results show that the internal organization of learned crystal latent representations can provide a useful control mechanism for generative modeling and support targeted exploration of materials chemical design space.

\bibliography{iclr2027_conference}
\bibliographystyle{iclr2027_conference}
\clearpage
\input{Appendix}

\end{document}

%% file: math_commands.tex
\usepackage{amsmath,amsfonts,bm}

\def\eqref#1{equation~\ref{#1}}

\def\1{\bm{1}}

\DeclareMathAlphabet{\mathsfit}{\encodingdefault}{\sfdefault}{m}{sl}
\SetMathAlphabet{\mathsfit}{bold}{\encodingdefault}{\sfdefault}{bx}{n}



%% file: Introduction.tex
\section{Introduction}

Crystalline materials discovery underpins advances in energy, electronics, catalysis, and many other technologies~\citep{butler2018machine,wang2026crystalline,cheng2026artificial}. Discovering such materials requires navigating an enormous chemical design space spanning diverse compositions, atomic arrangements, and lattice geometries. Conventional discovery workflows rely heavily on iterative experiments and computationally intensive first-principles calculations, limiting the scale and speed of materials exploration~\citep{potyrailo2011combinatorial,maier2019early,jain2011high}. To accelerate this process, deep generative models have emerged as a promising approach for learning distributions over crystal structures and directly proposing previously unseen candidates~\citep{xie2022crystal,luo2023towards,zhao2021high}. In particular, diffusion- and flow-based models have further advanced crystal generation by jointly modeling atomic species, fractional coordinates, and lattice parameters~\citep{xie2022crystal,jiao2023diffcsp,miller2024flowmm,zeni2025generative}.

Building on recent advances in latent generative modeling for molecules~\citep{xu2023geometric,wang2024gldm,bian2024hierarchical,ding2025scalable,zhang2025geometry,slautin2026integrating,luo2026towards}, recent work has explored latent generative models for improving crystal generation performance~\citep{khastagir2025crysldm,joshi2025allatom}. In a standard flow-based framework~\citep{joshi2025allatom}, a variational autoencoder (VAE) first learns continuous representations of crystals through encoding and reconstruction. A separate flow model then maps an initial source distribution to the distribution of VAE-encoded representations. During inference, the VAE decoder converts representations transported by the flow into crystal structures. Despite the central role of these representations in generation, existing work primarily focuses on the quality of generated crystals, with limited attention to the organization of the latent space itself. Understanding how these representations are organized could inform the design of latent generative models and sampling strategies.

We therefore ask: \emph{What organization emerges in the latent space of a crystal VAE, and how can it guide generation?} 
Recent work has shown that atomic representations learned by machine-learning interatomic potentials can reflect consistent geometric organization of atomic environments. In particular, representations of atoms from the same periodic-table group tend to cluster together~\citep{li2026platonic}. This motivates us to examine whether similar geometric organization can emerge when learning latent representations of crystals. To this end, we analyze a vanilla atom-wise crystal VAE trained only with reconstruction loss and Kullback--Leibler (KL) regularization. 

We find that its atomic latent representations naturally organize by elemental family, despite the absence of family labels during VAE training. Here, a family refers to a periodic-table category such as halogens or transition metals, following the mapping in Appendix~\ref{app:families}. As shown in Figure~\ref{fig:latent_semantics}, representations from the same family form coherent neighborhoods, while representations from different families occupy largely non-overlapping regions. We refer to this geometric organization as \emph{emergent chemical semantics}. This finding suggests that latent representations sampled from a family-specific region may preferentially decode into atoms from the corresponding family, providing a potential basis for controlling elemental family composition. However, the standard latent flow models~\citep{joshi2025allatom} do not make full use of such emergent chemical semantics. The challenge is therefore to determine how to incorporate it into latent flow modeling to guide generation.

To bridge this gap, we introduce \textbf{FamilySpace}, which transforms the observed semantics into a family-indexed multi-source distribution for latent flow matching. For each elemental family, FamilySpace constructs a local source component from the corresponding atomic representations in the pretrained VAE posterior.
A single shared flow is trained by pairing samples from each source component with target atomic latents from the same family, using the standard flow-matching objective. The VAE remains frozen, and family labels are used for source construction and source--target pairing. Figure~\ref{fig:familyspace_overview} illustrates the overall framework. At inference, a family template specifies the desired elemental families and their proportions, determining the source component for each atomic representation. Source selection therefore controls family composition while leaving exact element identities and crystal geometry to be generated by the model.

Experiments on MP-20 and MPTS-52 demonstrate competitive generation performance, while control experiments show that generated crystals accurately follow the requested family compositions. Moreover, selected chemically motivated templates increase the fraction of generated crystals predicted to be stable or metastable, without additional model optimization such as reinforcement learning. These results show how emergent chemical semantics can support both controllable generation and targeted materials exploration.

Our main contributions are as follows:
\begin{itemize}
    \item
    We identify \emph{emergent chemical semantics} in a vanilla atom-wise crystal VAE. Although trained only with reconstruction and KL regularization, the VAE learns latent representations that naturally organize by elemental family.

    \item
    We introduce FamilySpace, which converts this emergent chemical semantics into a family-indexed multi-source distribution for latent flow matching. Pairing source and target latents from the same family enables family-composition control through source selection.
    
    \item
    We demonstrate FamilySpace's precise family-composition control and competitive crystal generation performance. The controllable generation mechanism of FamilySpace enables targeted exploration of chemically meaningful regions of materials space,  
    without additional optimization techniques such as reinforcement learning.    

\end{itemize}

%% file: Related_work.tex
\section{Related Work}

{\bf Generative models for crystalline materials.} Recent crystal generative models predominantly model crystal variables directly, jointly generating atomic species, fractional coordinates, and lattice  parameters with diffusion or flow-based objectives~\citep{xie2022crystal,jiao2023diffcsp,miller2024flowmm,zeni2025generative}. Subsequent work has improved scalability, stability, conditioning, and exploration through stronger priors, symmetry-aware representations, property guidance, and multimodal inputs~\citep{sriram2024flowllm,cornet2025kinetic,kazeev2025wyckoff,kelvinius2025wyckoffdiff,hollmer2025open,das2025periodic,levy2025symmcd,chang2026space,khastagir2026llm,seong2026multimodal,hollmer2026open}.  These works have substantially advanced direct generative modeling of explicit crystal variables.

{\bf Latent generative modeling for crystalline materials.} Latent crystal generative models separate representation learning from generative distribution modeling.
CrysLDM performs diffusion over crystal representations learned by a VAE~\citep{khastagir2025crysldm}, while ADiT combines a shared all-atom VAE with latent flow matching to jointly model periodic crystals and non-periodic molecules~\citep{joshi2025allatom}.
In these models, the VAE learns to encode and reconstruct crystal structures, whereas the diffusion or flow model learns the distribution of the resulting latent representations.
In a related direction, Chemeleon learns aligned text and crystal representations for text-guided crystal generation~\citep{park2025exploration}.
Chemeleon2 further introduces a continuous VAE latent space, performing denoising diffusion in this space and incorporating reinforcement learning for materials exploration~\citep{park2026guiding}.

%% file: Preliminaries.tex
\section{Preliminaries}
\label{sec:preliminaries}


\subsection{Crystal representation and atom-wise latent space}
A periodic crystal containing $N$ atoms is represented as $\mathcal{M}=(\mathbf{A},\mathbf{F},\mathbf{L})$,
where $\mathbf{A}=(a_1,\ldots,a_N)$ denotes the atomic species, $\mathbf{F}\in[0,1)^{N\times3}$ contains the fractional coordinates, and $\mathbf{L}$ specifies the periodic unit cell. 
We parameterize $\mathbf{L}$ by its three cell lengths $\mathbf{l}=(a,b,c)$ and three cell angles $\boldsymbol{\alpha}=(\alpha,\beta,\gamma)$.
Following the latent representation proposed by \citet{joshi2025allatom}, we encode a crystal into an atom-wise latent sequence $\mathbf{Z} = (\mathbf{z}_1,\ldots,\mathbf{z}_N)\in\mathbb{R}^{N\times d}$, 
where each latent representation $\mathbf{z}_i$ is associated with one atom.

\subsection{Atom-wise VAE}
Following ADiT \citep{joshi2025allatom}, the first stage of latent generative modeling maps the crystal variables $\mathcal{M}$ into a continuous atom-wise latent space using a VAE. The encoder produces a Gaussian posterior
$q_{\phi}(\mathbf{z}_i\mid\mathcal{M})
=\mathcal{N}\!\left(\boldsymbol{\mu}_i,\operatorname{diag}(\boldsymbol{\sigma}_i^2)\right)$,
from which the atom-wise latent sequence $\mathbf{Z}$ is sampled. The decoder maps $\mathbf{Z}$ back to the crystal variables.
For periodic crystals, the reconstruction objective includes atomic species, fractional coordinates, cell lengths, and cell angles:
$\mathcal{L}_{\mathrm{rec}}
=\lambda_A \mathcal{L}_{A}+\lambda_F \mathcal{L}_{F}
+\lambda_{l} \mathcal{L}_{l}+\lambda_{\alpha} \mathcal{L}_{\alpha}$,
where $\mathcal{L}_{A}$ is the cross-entropy loss for atomic species, while $\mathcal{L}_{F}$, $\mathcal{L}_{l}$, and
$\mathcal{L}_{\alpha}$ are reconstruction losses for fractional coordinates, cell lengths, and cell angles, respectively. The complete VAE objective is
\begin{equation}
\mathcal{L}_{\mathrm{VAE}}
=
\mathcal{L}_{\mathrm{rec}}
+
\lambda_{\mathrm{KL}}
D_{\mathrm{KL}}
\left(
q_{\phi}(\mathbf{Z}\mid\mathcal{M})
\,\|\,\mathcal{N}(\mathbf{0},\mathbf{I})
\right).
\label{eq:vae_objective}
\end{equation}

The VAE is trained solely to reconstruct crystal structures and regularize the latent distribution. 

\subsection{Latent flow matching}
After training the VAE, the second stage of latent generative modeling learns to transform a simple source distribution into the distribution of encoded crystal latents. Following ADiT \citep{joshi2025allatom}, let $\mathbf{Z}^{(1)}$ denote a clean latent representation obtained from the frozen VAE encoder and let $\mathbf{Z}^{(0)}\sim p_0=\mathcal{N}(\mathbf{0},\mathbf{I})$
denote a sample from the standard Gaussian source. A linear conditional path between the source and target is defined as
\begin{equation}
    \mathbf{Z}^{(t)}
    =
    (1-t)\mathbf{Z}^{(0)}
    +
    t\mathbf{Z}^{(1)},
    \qquad
    t\sim\mathcal{U}(0,1).
\label{eq:linear_path}
\end{equation}
The corresponding conditional velocity field is 
\begin{equation}
\mathbf{u}_t\left(\mathbf{Z}^{(t)}\mid\mathbf{Z}^{(1)}\right)=\frac{\mathbf{Z}^{(1)}-\mathbf{Z}^{(t)}}{1-t}.
\end{equation}

ADiT parameterizes this vector field by predicting the clean endpoint from the intermediate latent,
$\widehat{\mathbf{Z}}^{(1)} =F_{\theta}\!\left(\mathbf{Z}^{(t)},t\right)$,
which induces the predicted velocity 
$\mathbf{v}_{\theta}(\mathbf{Z}^{(t)},t) =\frac{\widehat{\mathbf{Z}}^{(1)}-\mathbf{Z}^{(t)}}{1-t}$.
The objective minimizes the discrepancy between the predicted and target vector fields, as follows:
\begin{equation}
\mathcal{L}_{\mathrm{LFM}}
=
\mathbb{E}_{t,\mathbf{Z}^{(0)},\mathbf{Z}^{(1)}}
\left[
\frac{1}{(1-t)^2}
\left\|
\widehat{\mathbf{Z}}^{(1)}
-
\mathbf{Z}^{(1)}
\right\|_F^2
\right].
\label{eq:prelim_flow}
\end{equation}

At inference time, samples are initialized from $p_0$ and transported from $t=0$ to $t=1$ by integrating the learned vector field, after which the frozen VAE decoder reconstructs the terminal latent representation into a crystal.

%% file: Methodology.tex
\section{FamilySpace}
\label{sec:method}

\subsection{Emergent chemical semantics in a vanilla crystal VAE}
\label{sec:emergent_semantics}

\begin{figure*}[t]
    \centering
    \includegraphics[width=\textwidth]{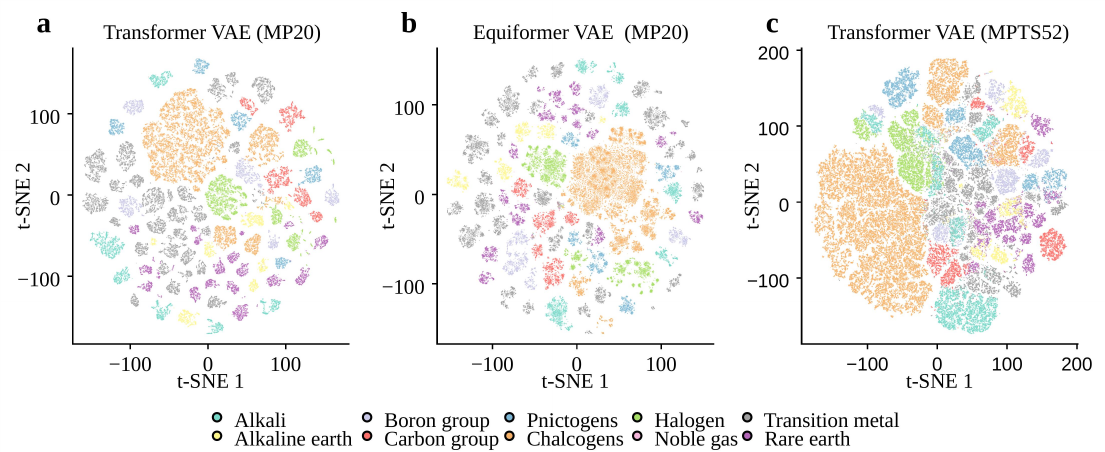}
\caption{
\textbf{Emergent chemical semantics in atom-wise latent representations.}
We visualize the posterior-mean atom-wise latent representations extracted from the complete test splits using t-SNE.
Each point represents an atom and is colored by its periodic-table family only for visualization.
(a) Transformer VAE on MP-20.
(b) Equiformer VAE on MP-20.
(c) Transformer VAE on MPTS-52.
}
    \label{fig:latent_semantics}
\end{figure*}


We begin with a vanilla atom-wise crystal VAE~\citep{joshi2025allatom} and examine how its learned atomic representations are organized in latent space. Given a crystal structure, the encoder produces an atom-wise posterior distribution for each atom, from which latent representations are sampled and passed to the decoder for crystal reconstruction.
The VAE is trained using only the standard reconstruction loss and KL regularization in Equation~\ref{eq:vae_objective}. 
We analyze the learned representations using the posterior mean $\boldsymbol{\mu}_i$ of each atom $i$.
As shown in Figure~\ref{fig:latent_semantics}, a consistent geometric organization emerges across both Transformer- and Equiformer-based VAEs.
Representations of atoms from the same elemental family form coherent neighborhoods, whereas representations from different families occupy largely non-overlapping regions. We refer to this spontaneously learned family-level organization as \emph{emergent chemical semantics}.
Further visualizations and analysis are provided in Appendix~\ref{app:representation_visualization}.
These findings suggest that reconstruction-based crystal representation learning naturally gives rise to family-level chemical organization in the latent space. In the following subsection, we investigate how this organization can be exploited to guide generation.

\begin{figure*}[t]
    \centering
    \includegraphics[width=\textwidth]{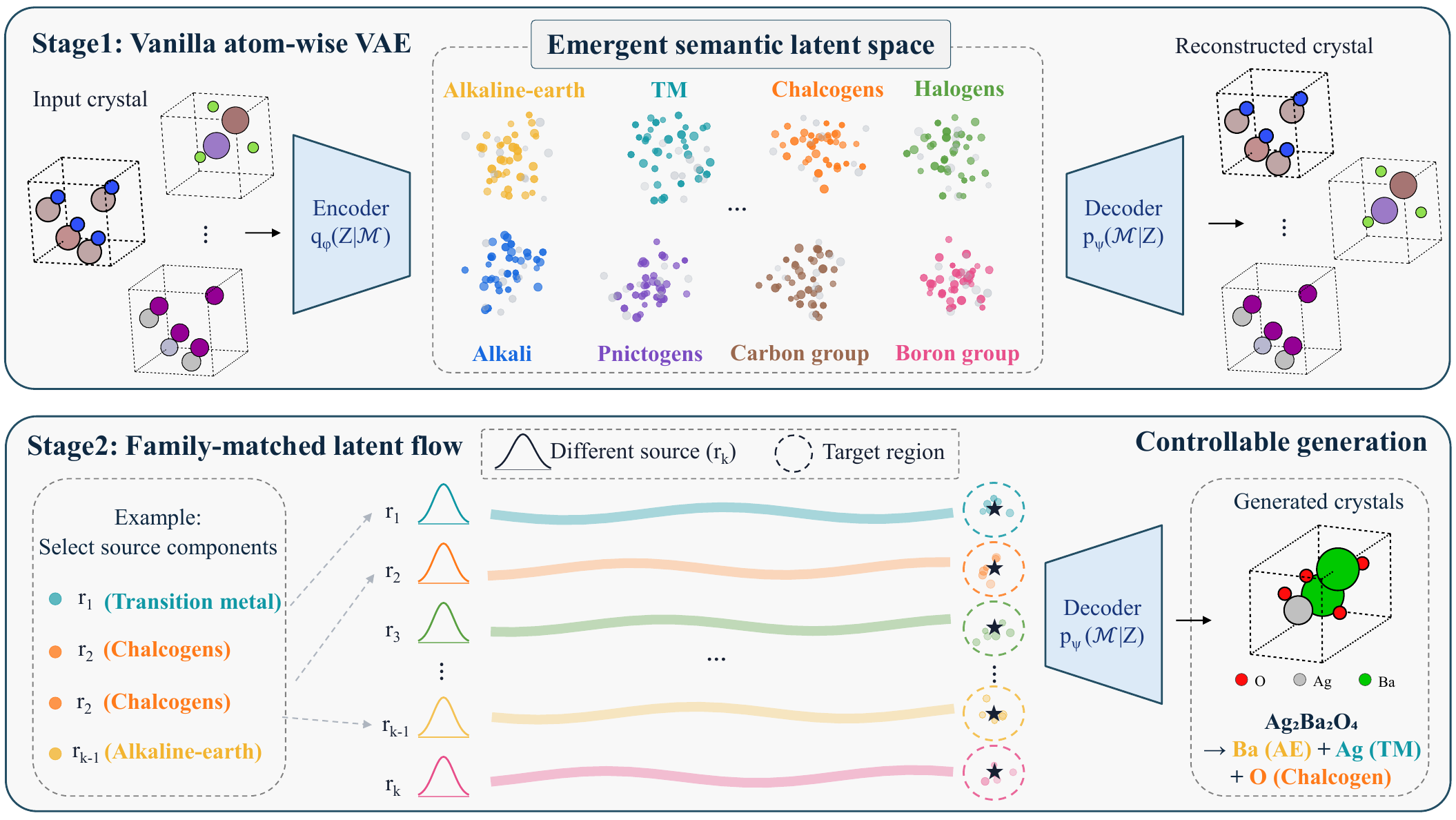}
    \caption{
    \textbf{Overview of FamilySpace.}
    \emph{Vanilla atom-wise VAE:} Reconstruction and KL regularization yield atom-wise representations with emergent family organization.
    \emph{Family-matched latent flow:} After the VAE is trained, source components are paired with posterior targets from the same family to train a shared flow model.
    \emph{Controllable generation:} a family template selects the source component for each atomic representation.
    The shared flow transports the latent representations, and the frozen VAE decoder generates atomic species and crystal geometry.
    }
    \label{fig:familyspace_overview}
\end{figure*}

\subsection{Family-matched latent flow}
\label{sec:familyspace_flow}

The emergent chemical semantics reveal that atom-wise latent representations from different elemental families tend to occupy largely non-overlapping regions of the VAE latent space.
This observation suggests a natural route to controllable crystal generation: if sampled latent representations can be associated with a particular family-specific region, they may preferentially decode into atoms from the corresponding elemental family.
To turn this observation into a generative mechanism, we first examine how source representations are associated with target families.

Let $p_1^{\mathrm{atom}}(\mathbf{z})$ denote the aggregated VAE posterior over atom-wise latent representations. Using family labels, it can be decomposed as
\begin{equation}
    p_1^{\mathrm{atom}}(\mathbf{z})
    =
    \sum_{k\in\mathcal{K}_{\mathcal{D}}}
    \pi_k p_1^{\mathrm{atom}}(\mathbf{z}\mid k),
    \label{eq:posterior_family_mixture}
\end{equation}
where $\mathcal{K}_{\mathcal{D}}$ is the set of families observed in the training data, $\pi_k$ is the corresponding weight, and $p_1^{\mathrm{atom}}(\mathbf{z}\mid k)$ is the family-conditional aggregated posterior.

In standard latent flow training, source representations are sampled from the same isotropic Gaussian regardless of family:
\begin{equation}
p_0^{\mathrm{atom}}(\mathbf{z}_0\mid k)
    =
    p_0^{\mathrm{atom}}(\mathbf{z}_0)
    =
    \mathcal{N}(\mathbf{z}_0;\mathbf{0},\mathbf{I}),
    k\in\mathcal{K}_{\mathcal{D}}.
\end{equation}
Thus, the source distribution of standard latent flow has no explicit correspondence with the family-specific regions of the VAE posterior, making this emergent semantics difficult to exploit directly for composition control.

To bridge this gap, FamilySpace introduces two changes to latent flow training: a family-indexed multi-source distribution and family-matched source--target coupling. Figure~\ref{fig:familyspace_overview} illustrates the overall framework. FamilySpace retains the pretrained VAE, the flow backbone, and the standard flow matching objective, while modifying only these two components. We describe each below.

\paragraph{Family-indexed multi-source distribution.}
FamilySpace replaces the Gaussian source distribution with a family-indexed multi-source distribution.
A deterministic mapping $g(a_i)$ assigns each atomic species to one of $K=10$ families, as specified in Appendix~\ref{app:families}.
For each observed family $k$, we select a representative posterior mean $\mathbf{m}_k$, referred to as the family medoid, and estimate its local variability from nearby same-family posterior means.
Perturbations around $\mathbf{m}_k$ are scaled according to this local variability and bounded using the distances between family medoids.
This construction defines a local source component $r_k$ for each family, with the components constrained to have non-overlapping supports when the medoids are distinct.
Details of source construction and sampling procedure are provided in Appendix~\ref{app:source_construction}, and source distribution ablation experiments are reported in Appendix~\ref{app:Source_ablation}.

The resulting atom-level source distribution during training is
\begin{equation}
    p_0^{\mathrm{atom}}(\mathbf{z})
    =
    \sum_{k\in\mathcal{K}_{\mathcal{D}}}
    \pi_k r_k(\mathbf{z}),
    \label{eq:family_source_mixture}
\end{equation}
where each component $r_k$ is indexed by an elemental family and $\pi_k$ is the weight for each component.
This multi-source distribution replaces only the source distribution used by the latent flow; the Gaussian prior used for VAE KL regularization remains unchanged.

\paragraph{Family-matched source-target coupling.}

A family-indexed source alone does not determine which family-specific region of the VAE posterior each source component should be transported toward. We therefore use family labels to construct family-matched source--target pairs during flow training. For a training crystal $\mathcal{M}=(\mathbf{A},\mathbf{F},\mathbf{L})$ with $N$ atoms, we get a target latent sequence from the frozen VAE posterior and select a source component for each atom according to its family:
\begin{equation}
    \mathbf{Z}^{(1)}
    \sim q_{\phi}(\mathbf{Z}\mid\mathcal{M}),
    \qquad
    \mathbf{z}^{(0)}_i
    \sim r_{g(a_i)},
    \qquad i=1,\ldots,N.
    \label{eq:family_matched_source}
\end{equation}
Source construction uses posterior means, whereas flow training uses posterior samples as targets.
Each source component is therefore paired only with targets from the same family:
\begin{equation}
    r_k(\mathbf{z}_0)
    \longleftrightarrow
    p_1^{\mathrm{atom}}(\mathbf{z}_1\mid k),
    \qquad k\in\mathcal{K}_{\mathcal{D}}.
    \label{eq:family_component_coupling}
\end{equation}

All target representations from a crystal remain together as a complete latent sequence.
A single shared flow is trained on the paired source and target sequences using the linear path and standard flow matching objective defined in Section~\ref{sec:preliminaries}.
Family labels determine the source components and training pairs, but are not provided as additional conditioning inputs to the flow network.
Method details are provided in Appendix~\ref{app:familyspace_implementation}.
Training details are provided in Appendix~\ref{app:algorithms}.

\subsection{Controllable generation}
\label{sec:familyspace_generation}

At inference, a requested family template represented as 
\begin{equation}
    \mathbf{f}=(f_1,\ldots,f_N),
    \qquad f_i\in\mathcal{K}_{\mathcal{D}},
    \label{eq:requested_family_template}
\end{equation}
specifies the desired family of each atomic representation.
The number of representations assigned to each family determines the requested family counts and ratios.
Given the template, source representations are sampled independently from the selected components:
\begin{equation}
    p_0\!\left(\mathbf{Z}^{(0)}\mid\mathbf{f}\right)
    =
    \prod_{i=1}^{N}
    r_{f_i}\!\left(\mathbf{z}^{(0)}_i\right).
    \label{eq:template_source}
\end{equation}
Thus, the template determines the family composition of the source sequence at inference.
The shared latent flow transports the complete latent sequence, and the frozen VAE decoder generates atomic species, fractional coordinates, and lattice parameters.

Control is implemented through source selection.
Each selected source component has been paired during training with targets from the requested family.
The template specifies family composition, while exact element identities and crystal geometry remain generated by the model.
Changing the template therefore directs generation toward a different family composition without further model training.
We evaluate how accurately the decoded compositions follow the requested templates in Section~\ref{sec:controllable_generation}.
Sampling details are provided in Appendix~\ref{app:sampling}.

%% file: Experiments.tex
\section{Experiments}
\label{sec:experiment}

\subsection{Experimental setup}
Our experiments evaluate the generation performance of FamilySpace and the accuracy of family-composition control through source selection. We further evaluate whether chemically motivated family templates can support application-oriented exploration.

\textbf{Datasets.}
We evaluate our method on the MP-20 dataset, which contains crystal structures with at most 20 atoms per primitive cell~\citep{xie2022crystal}, and the MPTS-52 dataset~\citep{baird2024matbench}, which consists of crystal structures with up to 52 atoms per unit cell. For both datasets, we follow the training, validation, and test splits adopted in prior work \citep{miller2024flowmm}. 


\textbf{Baselines.}
We compare FamilySpace with 
DiffCSP~\citep{jiao2023diffcsp},
FlowMM~\citep{miller2024flowmm},
MatterGen~\citep{zeni2025generative},
ADiT~\citep{joshi2025allatom},
SymmCD~\citep{levy2025symmcd},
SGEquiDiff~\citep{chang2026space}, CrystalDiT~\citep{yi2026crystaldit}, Chemeleon~\citep{park2025exploration}, and
Chemeleon2~\citep{park2026guiding} on MP-20.
On MPTS-52, we compare with ADiT and SGEquiDiff.

\begin{table*}[!t]
    \caption{
    Crystal generation results on MP-20. The best results are highlighted in bold, while the second-best results are underlined.
    }
    \label{tab:mp20_stability}
    \centering
    \small
    \setlength{\tabcolsep}{5.0pt}
    \renewcommand{\arraystretch}{1.08}

    \resizebox{\textwidth}{!}{%
    \begin{tabular}{lcccccccc}
        \toprule

        \multirow{2}{*}{Method}
        & \multirow{2}{*}{Valid (\%) $\uparrow$}
        & \multicolumn{3}{c}{Uniqueness / novelty (\%) $\uparrow$}
        & \multicolumn{2}{c}{Stability (\%) $\uparrow$}
        & \multicolumn{2}{c}{Metastability (\%) $\uparrow$} \\

        \cmidrule(lr){3-5}
        \cmidrule(lr){6-7}
        \cmidrule(lr){8-9}

        &
        & U
        & N
        & UN
        & S
        & SUN
        & MS
        & MSUN \\

        \midrule

        DiffCSP~\citep{jiao2023diffcsp}
        & 82.49
        & 96.54
        & 71.10
        & 70.79
        & 5.76
        & 1.55
        & 43.56
        & 18.66 \\

        FlowMM~\citep{miller2024flowmm}
        & 82.14
        & 97.86
        & 72.34
        & 72.20
        & 4.91
        & 1.40
        & 39.43
        & 15.80 \\

        MatterGen~\citep{zeni2025generative}
        & 83.58
        & \textbf{97.92}
        & \underline{74.96}
        & \textbf{74.68}
        & 4.96
        & 1.80
        & 44.74
        & 23.23 \\

        ADiT~\citep{joshi2025allatom}
        & 90.24
        & 83.90
        & 31.72
        & 30.81
        & 11.03
        & 1.68
        & \underline{77.91}
        & 13.39 \\

        SymmCD~\citep{levy2025symmcd}
        & 84.60
        & \underline{97.87}
        & 73.19
        & \underline{72.79}
        & 4.84
        & 1.53
        & 34.09
        & 10.90 \\

        SGEquiDiff~\citep{chang2026space}
        & 85.42
        & 96.56
        & 63.68
        & 63.36
        & 6.52
        & 1.46
        & 44.77
        & 12.36 \\

        CrystalDiT~\citep{yi2026crystaldit}
        & 84.30
        & 89.97
        & 38.60
        & 37.82
        & \underline{11.64}
        & \underline{2.65}
        & 75.69
        & 18.28 \\

        Chemeleon~\citep{park2025exploration}
        & 83.62
        & 97.15
        & 70.32
        & 69.71
        & 5.55
        & 1.62
        & 49.95
        & 21.82 \\

        Chemeleon2~\citep{park2026guiding}
        & 88.77
        & 75.43
        & 59.14
        & 50.43
        & 11.20
        & 2.19
        & 70.68
        & 28.51 \\

        \midrule
        \rowcolor{gray!15}
        FamilySpace (Training templates)
        & 89.48
        & 97.14
        & 69.63
        & 69.53
        & 6.26
        & 1.38
        & 42.99
        & 15.31 \\
        \rowcolor{gray!15}
        FamilySpace (Alkali--Halogen 1:1)
        & \textbf{99.85}
        & 47.74
        & 58.40
        & 45.25
        & \textbf{13.56}
        & 0.70
        & \textbf{90.23}
        & \underline{36.12} \\
        \rowcolor{gray!15}
        FamilySpace (RE--Halogen 1:3)
        & \underline{91.22}
        & 64.53
        & \textbf{92.58}
        & 63.00
        & 7.87
        & \textbf{4.97}
        & 70.36
        & \textbf{41.50} \\

        \bottomrule
    \end{tabular}%
    }
\end{table*}

\begin{table*}[!t]
    \caption{
    Crystal generation results on MPTS-52. The best results are highlighted in bold, while the second-best results are underlined.
    }
    \label{tab:mpts52_stability}
    \centering
    \small
    \setlength{\tabcolsep}{5.0pt}
    \renewcommand{\arraystretch}{1.08}

    \resizebox{\textwidth}{!}{%
    \begin{tabular}{lcccccccc}
        \toprule

        \multirow{2}{*}{Method}
        & \multirow{2}{*}{Valid (\%) $\uparrow$}
        & \multicolumn{3}{c}{Uniqueness / novelty (\%) $\uparrow$}
        & \multicolumn{2}{c}{Stability (\%) $\uparrow$}
        & \multicolumn{2}{c}{Metastability (\%) $\uparrow$} \\

        \cmidrule(lr){3-5}
        \cmidrule(lr){6-7}
        \cmidrule(lr){8-9}

        &
        & U
        & N
        & UN
        & S
        & SUN
        & MS
        & MSUN \\

        \midrule
        
        ADiT~\citep{joshi2025allatom}
        & 83.82
        & 93.42
        & 55.64
        & 55.15
        & \underline{9.77}
        & 2.32
        & 59.21
        & 18.33 \\
        
        SGEquiDiff~\citep{chang2026space}
        & 80.59
        & \textbf{97.69}
        & \underline{89.07}
        & \textbf{88.18}
        & 4.31
        & \underline{2.39}
        & 26.13
        & 16.05 \\

        \midrule
        \rowcolor{gray!15}
        FamilySpace (Training templates)
        & 78.59
        & \underline{96.75}
        & 84.47
        & \underline{83.73}
        & 5.32
        & 2.11
        & 31.60
        & 17.27 \\
        \rowcolor{gray!15}
        FamilySpace (Alkali--Halogen 1:1)
        & \underline{97.32}
        & 24.56
        & 39.65
        & 24.14
        & \textbf{18.75}
        & 0.63
        & \textbf{93.89}
        & \underline{21.10} \\
        \rowcolor{gray!15}
        FamilySpace (RE--Halogen 1:3)
        & \textbf{98.11}
        & 69.56
        & \textbf{95.22}
        & 69.28
        & 8.97
        & \textbf{5.10}
        & \underline{74.55}
        & \textbf{46.24} \\

        \bottomrule
    \end{tabular}%
    }
\end{table*}

\textbf{Implementation.}
Since FamilySpace is designed for controllable generation, it requires a family template specifying the requested family composition of the generated crystals. We therefore evaluate FamilySpace under different templates. For de novo generation, we consider three representative settings. The first is \emph{training-template sampling}, under which 10,000 samples are generated by drawing family compositions from the empirical training distribution. The other two are the \emph{Alkali--Halogen (1:1)} and \emph{RE--Halogen (1:3)} templates\footnote{For family combinations, AE denotes Alkaline earth, TM Transition metal, RE Rare earth, Ch Chalcogens, and Pn Pnictogens, following the family mapping in Appendix~\ref{app:families}.}, which specify the corresponding family ratios and are selected as chemically motivated targets for stability-oriented exploration. 
The template-selection procedure is in Appendix~\ref{app:targeted_protocols}.
Generation experiments use a separately trained Transformer VAE for each dataset and a single flow network shared across families within each dataset.
The VAE remains frozen during flow training and generation.
Architecture and training settings are provided in Appendices~\ref{app:vae_training} and~\ref{app:algorithms}, and sampling details are given in Appendix~\ref{app:sampling}.

\textbf{Evaluations.}
We evaluate generated structures in terms of validity, uniqueness and novelty, and thermodynamic stability. A crystal is considered valid only when it satisfies both compositional and structural validity criteria. Uniqueness (U) measures whether a generated crystal is distinct from other generated samples, while novelty (N) measures whether it is absent from the reference structure database; we further report their intersection $\mathrm{UN}=\mathrm{U}\cap\mathrm{N}$. Thermodynamic stability is evaluated after MatterSim relaxation (\texttt{MatterSim-v1.0.0-1M}) \citep{yang2024mattersim,zeni2025generative} using the energy above the convex hull, $E_{\mathrm{hull}}$. We define stable samples as $\mathrm{S}:E_{\mathrm{hull}}\leq0~\mathrm{eV/atom}$ and stable or metastable samples as
$\mathrm{MS}:E_{\mathrm{hull}}\leq0.1~\mathrm{eV/atom}$. Accordingly, we report $\mathrm{SUN}=\mathrm{S}\cap\mathrm{UN}$ and $\mathrm{MSUN}=\mathrm{MS}\cap\mathrm{UN}$ to quantify candidates that
are simultaneously thermodynamically favorable, unique, and novel.
More details are provided in Appendix~\ref{app:evaluation_details}.

\subsection{De novo generation}
\label{sec:De_novo}

Table~\ref{tab:mp20_stability} compares FamilySpace with baseline methods on the MP-20 dataset. Under the training-template sampling, FamilySpace achieves a substantially higher UN rate than ADiT, resulting in a higher MSUN rate. More importantly, the controllable generation capability of FamilySpace enables direct stability-oriented exploration of latent regions by simply selecting appropriate family templates, without requiring additional optimization techniques such as reinforcement learning.
With the Alkali--Halogen (1:1) template, the S and MS rates increase to 13.56\% and 90.23\%, respectively, exceeding all evaluated baselines. The RE--Halogen (1:3) template further achieves SUN and MSUN rates of 4.97\% and 41.50\%, respectively, again exceeding the corresponding baseline rates. 
Table~\ref{tab:mpts52_stability} compares FamilySpace with baseline methods on the MPTS-52 dataset. The results show trends consistent with those observed on MP-20.
These results demonstrate that FamilySpace can achieve strong generation performance.

\subsection{Controllable generation}
\label{sec:controllable_generation}
In this section, we evaluate whether selecting source components produces the requested family composition after latent transport and decoding.
All control metric definitions are provided in Appendix~\ref{app:control_metrics}.

\textbf{Single-family control.}
We first evaluate each source component individually.
For a requested family, all source representations are sampled from its component, transported by the shared flow, and decoded into a crystal.
We generate 1,000 samples for each of the ten families.
As shown in Table~\ref{tab:pure_region_control_main}, the average \emph{Target-family fraction} is 98.86\%, and the overall \emph{Family-count accuracy} is 95.20\%.
For nine families, the Target-family fraction exceeds 97.9\%.
The noble-gas family has the lowest Family-count accuracy (68.80\%) and contains only 38 training atoms.
Excluding noble gases, Family-count accuracy is 98.13\%.
These results show a strong empirical correspondence between source-component selection and decoded family composition.

\begin{table}[t]
    \caption{
    Single-family control on MP-20.
    \emph{Target-family fraction} measures the fraction of decoded atoms
    belonging to the requested family.
    \emph{Family-count accuracy} measures the percentage of samples whose
    decoded family-count vector exactly matches the requested one.
    Formal metric definitions are provided in Appendix~\ref{app:control_metrics}.
    }
    \label{tab:pure_region_control_main}
    \centering
    \small
    \setlength{\tabcolsep}{7pt}
    \begin{tabular}{@{}lrr@{}}
        \toprule
        Target family
        & Target-family fraction (\%)
        & Family-count accuracy (\%) \\
        \midrule
        Alkali            & 99.44 & 96.90 \\
        Alkaline earth    & 99.90 & 99.30 \\
        Boron group       & 99.97 & 99.80 \\
        Carbon group      & 99.93 & 99.70 \\
        Pnictogens        & 99.98 & 99.70 \\
        Chalcogens        & 99.92 & 99.50 \\
        Halogen           & 99.87 & 99.20 \\
        Noble gas         & 91.78 & 68.80 \\
        Transition metal  & 99.90 & 99.50 \\
        Rare earth        & 97.92 & 89.60 \\
        \midrule
        All families      & 98.86 & 95.20 \\
        \bottomrule
    \end{tabular}
\end{table}

\textbf{Multi-family ratio control.}
We next combine source components to generate crystals with prescribed family ratios.
We evaluate five family combinations containing two to four families, with 10,000 samples per combination.
For example, $1{:}1{:}2$ and $2{:}2{:}4$ represent the same reduced ratio.
As shown in Table~\ref{tab:family_control}, FamilySpace achieves 99.992\% Family-set accuracy and 99.968\% Family-ratio accuracy across all 50,000 samples.
These results show that source components can be combined to control the requested family composition with high accuracy.
In addition, family-level control does not fix the exact elements. The corresponding diversity results are reported in Appendix~\ref{app:control_diversity}.

\noindent\textbf{Comparison with Other Composition-Conditioned Generative Models.}
We compare FamilySpace with MatterGen~\citep{zeni2025generative} and Chemeleon~\citep{park2025exploration} on two family templates from MP-20: \emph{Alkali--Halogen (1:1)} and \emph{RE--Halogen (1:3)}. Each method generates 10,000 samples per template.
Since MatterGen and Chemeleon do not directly use family templates, we construct suitable conditions for these models based on the samples generated by FamilySpace. For MatterGen, which controls the chemical system (e.g., Li--Co--O) through classifier-free guidance (CFG), we use the official \texttt{chemical\_system} checkpoint with a CFG scale of 2. We restrict the chemical systems to those observed in the corresponding MP-20 training templates (52 for RE--Halogen and 41 for Alkali--Halogen). The number of samples generated for each chemical system is determined by its frequency among the 10,000 FamilySpace samples.
For Chemeleon, we use the official checkpoint with a CFG scale of 2. We construct the conditions based on the 10,000 FamilySpace samples. Specifically, we provide the reduced chemical formula and the total number of atoms for each generated structure. For example, if a FamilySpace output contains $\mathrm{La_2Cl_6}$, we use $\mathrm{LaCl_3}$ as the composition condition and set the number of atoms to 8.
We also compare FamilySpace with the crystal structure prediction (CSP) model OMatG \citep{hollmer2025open}. We use its official CSP checkpoint to generate crystal structures conditioned on the chemical formulas of the 10,000 FamilySpace samples. We then evaluate the structures using the same generation metrics.

As shown in Table~\ref{tab:composition_conditioned}, FamilySpace achieves the highest MSUN rate on both templates. In addition to these metrics, we evaluate whether the generated samples satisfy the specified family constraints. As shown in Table~\ref{tab:composition_conditioned_control}, almost all samples generated by FamilySpace contain only elements from the specified families, demonstrating accurate control over element families. MatterGen and Chemeleon achieve lower family-set accuracy than FamilySpace, indicating that their outputs less consistently contain exactly the requested elemental families. MatterGen only controls the chemical system without specifying the element ratios, resulting in relatively low Family-ratio accuracy. Although Chemeleon uses text conditioning with CFG to specify the RE:Halogen ratio of 1:3, its Family-ratio accuracy remains considerably lower than that of FamilySpace. These results show that FamilySpace provides accurate control under the evaluated conditioning settings.

\begin{table*}[t]
    \caption{
        Multi-family ratio control on MP-20.
        \emph{Family-set accuracy} measures whether the generated crystal contains exactly the requested families, while
        \emph{Family-ratio accuracy} additionally requires the reduced family ratio to match the request.
        Formal metric definitions are provided in Appendix~\ref{app:control_metrics}.
    }
    \label{tab:family_control}
    \centering
    \small
    \setlength{\tabcolsep}{5pt}
    \renewcommand{\arraystretch}{1.08}

    \begin{tabular}{llcc}       
        \toprule
        Family combination
        & Tested ratios
        & Family-set accuracy (\%)
        & Family-ratio accuracy (\%) \\
        \midrule

        TM--Ch
        & $1{:}1$, $1{:}2$, $2{:}3$, $3{:}4$
        & 99.99 & 99.94 \\

        Alkali--TM--Ch
        & $1{:}1{:}2$, $1{:}2{:}4$, $1{:}1{:}3$, $2{:}1{:}3$
        & 99.99 & 99.95 \\

        AE--TM--Ch
        & $1{:}1{:}3$, $1{:}2{:}4$, $1{:}1{:}2$
        & 100.00 & 99.98 \\

        RE--TM--Ch
        & $1{:}1{:}3$, $1{:}1{:}4$, $2{:}1{:}4$
        & 99.99 & 99.98 \\

        Alkali--TM--Pn--Ch
        & $1{:}1{:}1{:}4$, $1{:}1{:}2{:}7$, $2{:}1{:}1{:}4$
        & 99.99 & 99.99 \\

        \midrule
        All 50,000 samples
        & & 99.992
        & 99.968 \\
        \bottomrule
    \end{tabular}
\end{table*}

\begin{table*}[!t]
    \caption{Comparison with other composition-conditioned generative models.}
    \label{tab:composition_conditioned}
    \centering
    \small
    \setlength{\tabcolsep}{5.0pt}
    \renewcommand{\arraystretch}{1.08}
    \resizebox{\textwidth}{!}
    {%
    \begin{tabular}{lcccccccc}    
        \toprule
        \multirow{2}{*}{Method}
        & \multirow{2}{*}{Valid (\%) $\uparrow$}
        & \multicolumn{3}{c}{Uniqueness / novelty (\%) $\uparrow$}
        & \multicolumn{2}{c}{Stability (\%) $\uparrow$}
        & \multicolumn{2}{c}{Metastability (\%) $\uparrow$}
        \\
        \cmidrule(lr){3-5}\cmidrule(lr){6-7}\cmidrule(lr){8-9}
        & & U & N & UN & S & SUN & MS & MSUN \\
        \midrule
        \multicolumn{9}{@{}l}{\textit{RE--Halogen ($1{:}3$)}} \\
        MatterGen~\citep{zeni2025generative} & 90.09 & 23.11 & 47.79 & 21.11 & \underline{11.51} & 2.17 & \textbf{86.26} & 13.00  \\
        Chemeleon~\citep{park2025exploration} & 84.77 & \underline{29.52} & \underline{67.02} & \underline{27.94} & \textbf{13.26} & \underline{3.77} & 71.44 & \underline{15.16} \\
        OMatG (CSP)~\citep{hollmer2025open} & \textbf{100} & 8.14 & 34.72 & 7.12 & 9.12 & 0.77 & \underline{81.87} & 5.04  \\     
        \rowcolor{gray!15}
        FamilySpace & \underline{91.22} & \textbf{64.53} & \textbf{92.58} & \textbf{63.00} & 7.87 & \textbf{4.97} & 70.36 & \textbf{41.50}  \\        
        \midrule
        \multicolumn{9}{@{}l}{\textit{Alkali--Halogen ($1{:}1$)}} \\
        MatterGen~\citep{zeni2025generative} & 73.82 & 40.25 & 51.95 & 36.93 & \textbf{17.44} & \textbf{4.48} & 83.68 & \underline{29.18} \\
        Chemeleon~\citep{park2025exploration} & 82.39 & \underline{43.91} & \underline{58.19} & \underline{41.46} & 14.20 & \underline{2.37} & 79.63 & 28.29  \\
        OMatG (CSP)~\citep{hollmer2025open} & \textbf{99.91} & 22.88 & 40.44 & 20.65 & \underline{17.32} & 0.89 & \textbf{94.81} & 17.52 \\    
        \rowcolor{gray!15}
        FamilySpace & \underline{99.85} & \textbf{47.74} & \textbf{58.40} & \textbf{45.25} & 13.56 & 0.70 & \underline{90.23} & \textbf{36.12} \\        
        \bottomrule
    \end{tabular}%
    }
\end{table*}

\begin{table*}[!t]
    \caption{Comparison of composition control accuracy on MP-20.}
    \label{tab:composition_conditioned_control}
    \centering
    \small
    \setlength{\tabcolsep}{7.0pt}
    \renewcommand{\arraystretch}{1.08}
    \begin{tabular}{lrr}    
        \toprule
        Method & Family-set accuracy (\%) $\uparrow$ & Family-ratio accuracy (\%) $\uparrow$\\
        \midrule
        \multicolumn{3}{@{}l}{\textit{RE--Halogen ($1{:}3$)}} \\
        MatterGen~\citep{zeni2025generative} & 77.02 & 66.91 \\
        Chemeleon~\citep{park2025exploration} & \underline{92.79} & \underline{81.90} \\
        \rowcolor{gray!15}
        FamilySpace & \textbf{99.66} & \textbf{99.66} \\        
        \midrule
        \multicolumn{3}{@{}l}{\textit{Alkali--Halogen ($1{:}1$)}} \\
        MatterGen~\citep{zeni2025generative} & 97.89 & 68.88 \\
        Chemeleon~\citep{park2025exploration} & \underline{98.99} & \underline{81.34} \\
        \rowcolor{gray!15}
        FamilySpace & \textbf{100.00} & \textbf{99.92} \\        
        \bottomrule
    \end{tabular}
\end{table*}

\textbf{Application-oriented exploration.}
Beyond reproducing prescribed family compositions, FamilySpace allows generation to be directed toward chemically motivated regions of materials space. We examine whether this capability enables access to compositions
associated with reported functional materials. We consider the five family combinations in Table~\ref{tab:family_control}, generate 10,000 structures for each combination, and retain candidates satisfying the MSUN criterion. Detailed screening results are reported in Appendix~\ref{app:application_screening}.

We conduct a composition-level literature comparison to identify generated candidates whose compositions have been investigated for specific applications in recent studies. Representative examples span several family templates.
The Alkali--TM--Ch template produces $\mathrm{NaCrO_2}$, which has been investigated as a sodium-ion battery cathode, with improved rate capability and cycling stability achieved through nitrogen-doped carbon coating~\citep{cai2025nitrogen}. The AE--TM--Ch template yields $\mathrm{SrTiO_3}$, whose reported single-crystal form has been investigated for photocatalytic overall water splitting~\citep{luo2025intrafacet}, and $\mathrm{BaZrO_3}$, a perovskite host relevant to proton-conducting ceramic electrolytes~\citep{tang2025phase}.
The RE--TM--Ch template yields $\mathrm{LaMnO_3}$, a composition investigated for oxygen-evolution electrocatalysis in the form of unconventional cubic nanosheets~\citep{dai2025microwave}.
The Alkali--TM--Pn--Ch template produces $\mathrm{Li_2MnFe(PO_4)_2}$, which is stoichiometrically equivalent to $\mathrm{LiMn_{0.5}Fe_{0.5}PO_4}$, a composition studied as a lithium-ion battery cathode with engineered carbon coatings~\citep{hou2025precursor}. These examples show that family-controlled generation can access compositions associated with energy storage, energy conversion, and catalysis. Visualizations of representative generated structures are provided in Appendix~\ref{app:application_screening}.

This analysis provides a preliminary assessment of the application relevance of the generated compositions. Composition-level correspondence does not establish equivalence to the reported crystal phases or functional performance, which may also depend on doping, coatings, surface structure, and morphology. DFT calculations and experimental validation are beyond the scope of this study. Further comparison with reference crystal structures, property calculations, and experimental characterization would be needed to assess the practical potential of the generated candidates.

%% file: Appendix.tex
\section*{AI use statement}
In this work, we used generative AI tools to assist with language polishing, experimental code development, and literature search. All AI-assisted text was reviewed by the authors. All AI-assisted code was tested and verified to work as expected. Relevant literature identified through AI-assisted searches was verified against the original sources.

\appendix

\section*{Appendix}

\section{VAE Details}
\label{app:representation_details}

\subsection{Periodic-table family mapping}
\label{app:families}

Let $\mathcal{K}=\{0,\ldots,K-1\}$, with $K=10$, denote the fixed family vocabulary. Table~\ref{tab:family-map} specifies the lookup $g:a\mapsto k$ used throughout the paper. Family labels are introduced \emph{after} VAE training for representation analysis, source construction, and source--target coupling. At inference, a requested family template selects from $\mathcal{K}_{\mathcal{D}}\subseteq\mathcal{K}$, the families observed in the training data. 

\begin{table}[!htbp]
    \caption{The deterministic family lookup used by FamilySpace. The category names denote the implemented partition: hydrogen is assigned to Alkali, and Rare earth includes both lanthanides and actinides. These assignments are modeling conventions.}
    \label{tab:family-map}
    \centering
    \small
    \setlength{\tabcolsep}{5pt}
    \begin{tabularx}{\linewidth}{@{}clX@{}}
        \toprule
        ID & Family & Elements \\
        \midrule
        0 & Alkali & H, Li, Na, K, Rb, Cs, Fr \\
        1 & Alkaline earth (AE) & Be, Mg, Ca, Sr, Ba, Ra \\
        2 & Boron group & B, Al, Ga, In, Tl, Nh \\
        3 & Carbon group & C, Si, Ge, Sn, Pb, Fl \\
        4 & Pnictogens (Pn)& N, P, As, Sb, Bi, Mc \\
        5 & Chalcogens (Ch) & O, S, Se, Te, Po, Lv \\
        6 & Halogen & F, Cl, Br, I, At, Ts \\
        7 & Noble gas & He, Ne, Ar, Kr, Xe, Rn, Og \\
        8 & Transition metal (TM) & Groups 3--12, excluding lanthanides and actinides \\
        9 & Rare earth (RE) & Lanthanides and actinides \\
        \bottomrule
    \end{tabularx}
\end{table}

\subsection{Vanilla atom-wise VAE}
\label{app:vae_training}
\label{app:mpts52_training}

Both datasets use an atom-wise Transformer VAE with an eight-dimensional latent for each atom. The encoder and decoder each have eight layers, hidden width 512, eight attention heads, and feed-forward width 2,048. The encoder predicts a diagonal Gaussian posterior, and the decoder reconstructs atomic species, fractional coordinates, cell lengths, and cell angles. The architecture is shared across datasets, but the model parameters are learned separately using dataset-specific reconstruction objectives.

\paragraph{MP-20.}
The VAE uses
\begin{equation}
    \mathcal{L}_{\mathrm{VAE}}^{\mathrm{MP20}}
    =2\mathcal{L}_{A}+5\mathcal{L}_{F}^{\mathrm{PBC}}
     +\mathcal{L}_{l}+10\mathcal{L}_{\alpha}
     +10^{-5}\mathcal{L}_{\mathrm{KL}}.
    \label{eq:appendix_vae_loss}
\end{equation}
Atomic species use cross-entropy. The coordinate residual is the componentwise minimum-image displacement in fractional space:
\begin{equation}
    \Delta\mathbf{f}_i
    =\operatorname{remainder}
      \!\left(\widehat{\mathbf{f}}_i-\mathbf{f}_i+\tfrac12,1\right)
      -\tfrac12,
    \qquad
    \ell_F^{\mathrm{PBC}}(i)=\tfrac13\|\Delta\mathbf{f}_i\|_2^2.
    \label{eq:pbc_fractional_loss}
\end{equation}
The factor $1/3$ follows the implementation's mean over coordinate components. This residual does not weight fractional displacements by the cell metric. The VAE is trained for 2,000 epochs.

\paragraph{MPTS-52.}
The MPTS-52 VAE uses ordinary squared error on fractional coordinates:
\begin{equation}
    \ell_F^{\mathrm{ordinary}}(i)
    =\tfrac13\|\widehat{\mathbf{f}}_i-\mathbf{f}_i\|_2^2,
    \qquad
    \mathcal{L}_{\mathrm{VAE}}^{\mathrm{MPTS52}}
    =\mathcal{L}_{A}+10\mathcal{L}_{F}^{\mathrm{ordinary}}
     +\mathcal{L}_{l}+10\mathcal{L}_{\alpha}
     +10^{-5}\mathcal{L}_{\mathrm{KL}}.
    \label{eq:mpts52_vae_loss}
\end{equation}
The training run was configured for 2,000 epochs.  All MPTS-52 generation results use this same frozen VAE.

\paragraph{Optimization and augmentation.}
Coordinate and atom-type losses are averaged over atoms in the batch. Length and angle losses first average over their three components and then over structures. The KL term sums over the eight latent dimensions and averages over atoms. 

Training applies a random global translation and a common rotation of the Cartesian positions and cell. Independently, 10\% of atomic-species inputs are masked and 10\% of Cartesian positions receive Gaussian noise with standard deviation $0.1~\text{\AA}$ per component. Fractional coordinates are recomputed and wrapped into the unit cell. Targets retain the global augmentation but exclude the masking and added noise. Both runs use AdamW with learning rate $10^{-4}$, zero weight decay, $(\beta_1,\beta_2)=(0.9,0.999)$, and $\epsilon=10^{-8}$. There is no learning-rate schedule. 

\begin{table}[!htbp]
    \caption{VAE architecture and optimization settings. }
    \label{tab:vae_hyperparameters}
    \centering
    \small
    \setlength{\tabcolsep}{4pt}
    \begin{tabularx}{\linewidth}{@{}Xll@{}}
        \toprule
        Setting & MP-20 & MPTS-52 \\
        \midrule
        Encoder / decoder layers & 8 / 8 & 8 / 8 \\
        Hidden / latent width & 512 / 8 & 512 / 8 \\
        Attention heads / feed-forward width & 8 / 2048 & 8 / 2048 \\
        Activation / normalization & GELU / pre-norm & GELU / pre-norm \\
        Dropout & 0 & 0 \\
        Optimizer / learning rate & AdamW / $10^{-4}$ & AdamW / $10^{-4}$ \\
        Weight decay / learning-rate scheduler & 0 / none & 0 / none \\
        Batch size / Configured epochs & 128 / 2000 & 256 / 2000 \\
        Gradient-norm clipping  & 10  & 10  \\
        \bottomrule
    \end{tabularx}
\end{table}

\subsection{Representation extraction and visualization}
\label{app:representation_visualization}

\textbf{Representation extraction.}
We extract the posterior mean of every atom in the test split, obtaining 93,158 representations from 9,046 crystals for MP-20 and 202,522 representations from 8,096 crystals for MPTS-52. Each dataset is encoded using its corresponding frozen VAE.
The posterior means, rather than stochastic posterior samples, provide the deterministic atom-wise representations used for visualization. For both datasets, t-SNE is applied with perplexity 40 and 1500
iterations.

\begin{figure*}[h]
    \centering
    \includegraphics[width=\textwidth]{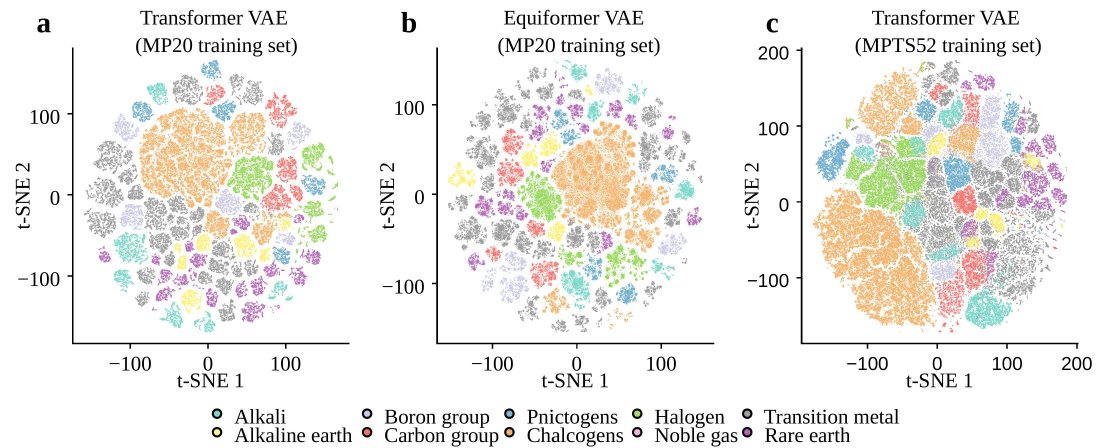}
    \caption{
    \textbf{Emergent chemical semantics in atom-wise latent representations on the training sets.}
We visualize the posterior-mean atom-wise latent representations extracted from the complete training splits using t-SNE.
    (a) Transformer VAE on MP-20.
    (b) Equiformer VAE on MP-20.
    (c) Transformer VAE on MPTS-52.
    All representations are colored by periodic-table family.
    }
    \label{fig:latent_semantics_train}
\end{figure*}

We additionally visualize the posterior representations extracted from the complete training splits using the same procedure. The MP-20 training split contains 282,847 atoms from 27,136 crystals, while the MPTS-52 training split contains 501,334 atoms from 27,380 crystals. As shown in Figure~\ref{fig:latent_semantics_train}, the family-level
organization observed in the test representations is also clearly present in the training representations.

\begin{figure*}[h]
    \centering
    \includegraphics[width=\textwidth]{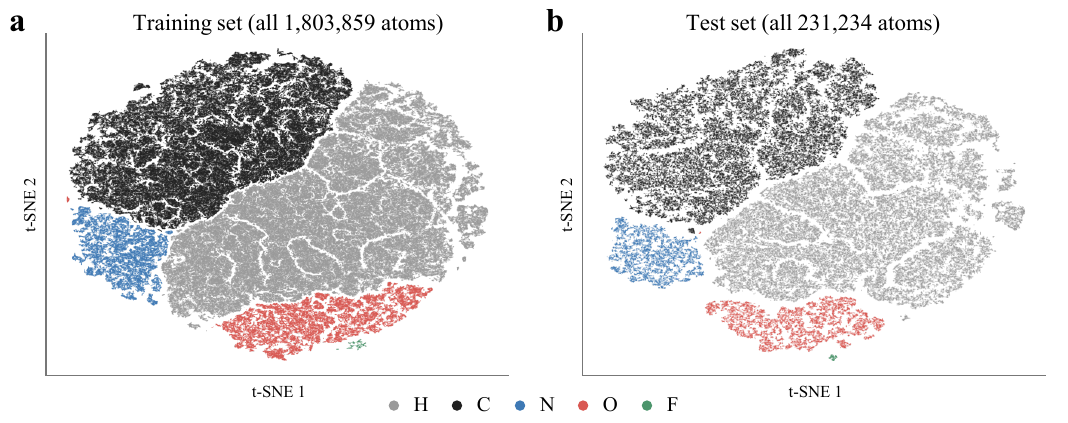}
    \caption{
\textbf{Emergent chemical semantics in atom-wise latent representations on QM9.}
We visualize the posterior-mean atom-wise latent representations extracted from the complete
(a) training split and (b) test split, respectively.
    }
    \label{fig:qm9_latent_semantics_train}
\end{figure*}

Because the ADiT VAE architecture is also applicable to molecular generation, we further visualize the latent space learned by a VAE trained on the QM9 molecular dataset.
As shown in Figure~\ref{fig:qm9_latent_semantics_train}, semantically meaningful organization also emerges in the latent space of the VAE trained on QM9.
This observation suggests that emergent semantic organization may be a more general phenomenon in atom-wise latent spaces for atomic systems.
In this work, we mainly focus on crystal generation; extending this investigation to molecules and other atomistic systems is an interesting direction for future work.

\textbf{Quantitative family-neighborhood analysis.}
To complement the qualitative results in
Figure~\ref{fig:latent_semantics}, we further quantify the local family-level organization of the Transformer VAE posterior means on MP-20 and MPTS-52. We evaluate the posterior mean representation using all atoms in the test split, including 93,158 atoms for MP-20 and 202,522 atoms for MPTS-52. 
Nearest neighbors are identified using exact Euclidean distance in the posterior space. 
We report two nearest-neighbor metrics.
\emph{Family 1-NN} measures the percentage of atoms whose nearest neighbor belongs to the same periodic-table family.
\emph{Family 5-NN} measures the percentage of atoms for which all five nearest neighbors belong to the same periodic-table family as the query atom.
As shown in Table~\ref{tab:posterior_family_knn}, the posterior representations exhibit strong local family-level organization on both datasets. On MP-20, all well-represented families achieve above 95\% accuracy for both nearest-neighbor metrics. MPTS-52 shows slightly
weaker but still strong local organization.

\begin{table*}[t]
\caption{
Quantitative family-neighborhood analysis of Transformer VAE posterior
means on the MP-20 and MPTS-52 test sets.
$N$ denotes the number of test atoms in each family.
}
\label{tab:posterior_family_knn}
\centering
\small
\setlength{\tabcolsep}{5.5pt}
\renewcommand{\arraystretch}{1.08}
\resizebox{\textwidth}{!}{%
\begin{tabular}{@{}lrrr|rrr@{}}
    \toprule
    \multirow{2}{*}{Family}
    & \multicolumn{3}{c}{MP-20}
    & \multicolumn{3}{c}{MPTS-52} \\
    \cmidrule(lr){2-4}
    \cmidrule(lr){5-7}
    & $N$
    & Family 1-NN (\%)
    & Family 5-NN (\%)
    & $N$
    & Family 1-NN (\%)
    & Family 5-NN (\%) \\
    \midrule

    Alkali
    & 6,647  & 99.8195 & 98.9619
    & 20,206 & 98.0006 & 90.2108 \\

    Alkaline earth
    & 4,322  & 99.6992 & 98.9125
    & 8,382  & 97.3157 & 87.5447 \\

    Boron group
    & 5,724  & 99.9651 & 99.8078
    & 11,075 & 96.2799 & 87.0429 \\

    Carbon group
    & 5,754  & 99.9305 & 99.5829
    & 12,791 & 97.9595 & 90.8060 \\

    Pnictogens
    & 4,848  & 99.9587 & 99.6700
    & 15,444 & 98.9964 & 94.3603 \\

    Chalcogens
    & 23,824 & 99.9706 & 99.8531
    & 79,150 & 99.7688 & 98.8932 \\

    Halogen
    & 8,598  & 99.4766 & 96.9877
    & 14,282 & 98.1725 & 91.3387 \\

    Noble gas
    & 11     & 36.3636 & 0.0000
    & 81     & 40.7407 & 0.0000 \\

    Transition metal
    & 23,997 & 99.6666 & 98.4665
    & 26,765 & 95.7183 & 83.6279 \\

    Rare earth
    & 9,433  & 99.6396 & 98.1024
    & 14,346 & 96.8144 & 86.4213 \\

    \midrule
    All atoms
    & 93,158
    & 99.78
    & 98.91
    & 202,522
    & 98.25
    & 92.58 \\

    \bottomrule
\end{tabular}%
}
\end{table*}

\begin{figure}[h]
    \centering
    \includegraphics[width=\linewidth]{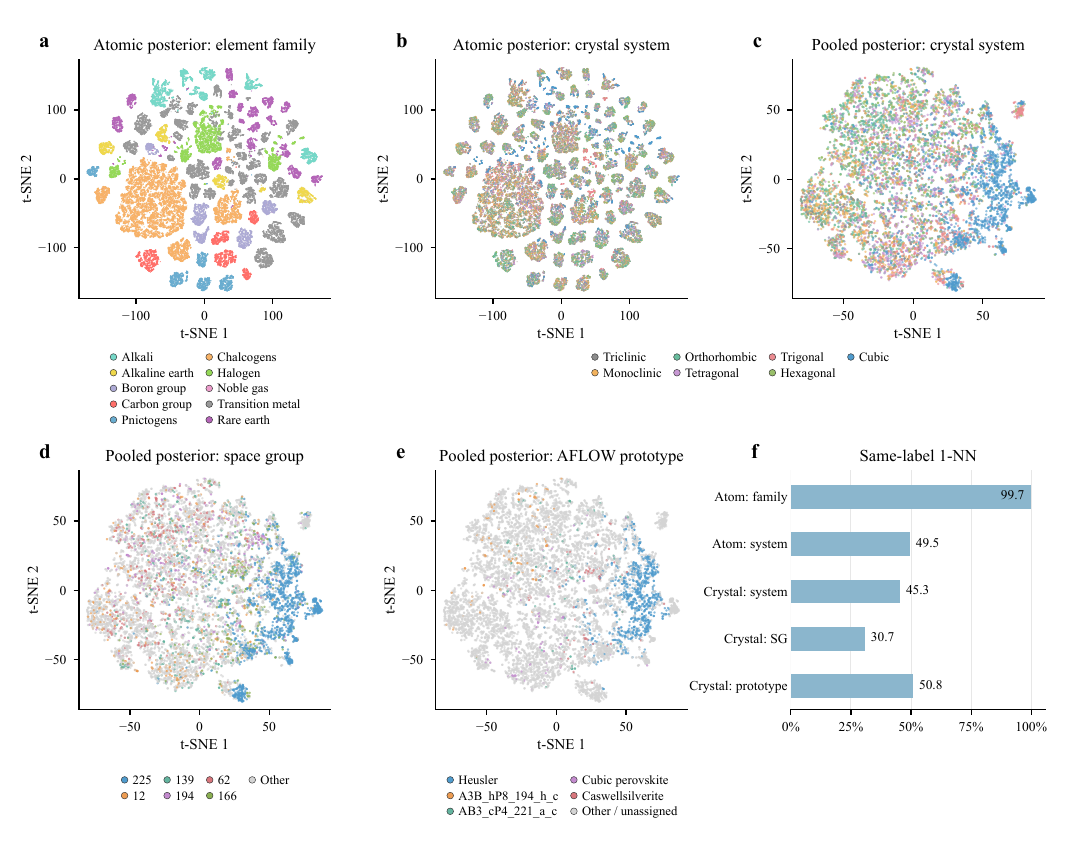}
\caption{
\textbf{Structural analysis in the VAE latent space.}
\textbf{a,b}, The same atom-level t-SNE embedding colored by elemental family and crystal system, respectively.
\textbf{c--e}, The same crystal-level t-SNE embedding colored by crystal system, space group, and AFLOW prototype, respectively.
\textbf{f}, Same-label 1-NN metric.
}
    \label{fig:mp20_vae_structural_semantics}
\end{figure}

\textbf{Structural analysis in the VAE latent space}.
We also examine whether the VAE captures information about crystal systems, space groups, and structural prototypes.
We use the 8-dimensional posterior mean vectors of 93,158 atoms from 9,046 MP-20 test crystals.
For atom-level analysis, we use individual atom vectors, while for crystal-level analysis, we average the atom vectors within each crystal. Structural prototype labels are assigned by matching the structures to the AFLOW prototype library.
Figure~\ref{fig:mp20_vae_structural_semantics} shows the t-SNE visualizations of these representations. Figures~\ref{fig:mp20_vae_structural_semantics}a,b show the same atom-level embedding, colored by elemental family and crystal system, respectively. Figures~\ref{fig:mp20_vae_structural_semantics}c--e show the same crystal-level embedding, colored by crystal system, space group, and structural prototype, respectively. 
Elemental families form clear clusters, while different structural labels show considerable overlap. 
To quantify these observations, we calculate \emph{same-label 1-NN}, similar to the previously used \emph{Family 1-NN} metric. It measures the percentage of samples whose nearest neighbors share the same label. For atom-level comparisons, we exclude all atoms from the same crystal. As shown in Figure~\ref{fig:mp20_vae_structural_semantics}f, elemental families show much stronger agreement than structural labels. 
For structural prototypes, we use pymatgen to match crystals to the AFLOW prototype library without requiring identical element types~\citep{ong2013python, mehl2017aflow,curtarolo2012aflow}. Among the 9,046 test crystals, 2,567 can be uniquely matched to 143 different prototypes. These representations do not form clear clusters in the latent space.
Overall, the VAE latent space shows much clearer clustering by elemental family than by structural labels.

\FloatBarrier
\section{FamilySpace Details}
\label{app:familyspace_implementation}

FamilySpace is constructed after the VAE has been trained and frozen. The VAE retains its standard Gaussian prior for KL regularization; the family-indexed multi-source distribution replaces only the source distribution of the subsequent latent flow. This section specifies source construction, training, and sampling.

\subsection{Source construction}
\label{app:source_construction}
Our goal is to construct a local source distribution for each periodic-table family using the posterior representations of the frozen VAE. Each source component is specified by three quantities: a representative center $\mathbf{m}_k$, a scale $\mathbf{s}_k$, and a support radius $\rho_k$. We first select an observed posterior representation near the center of each family, then estimate the local variation around it using nearby same-family representations. We use these estimates to sample perturbations around the representative and bound their magnitude according to the distance to neighboring family centers. Together, these steps produce the family-indexed source components.

\paragraph{Family representative center estimation.}
Let $\boldsymbol{\mu}_i\in\mathbb{R}^{d}$ denote the posterior mean of training atom $i$, and let $\boldsymbol{\mu}_{\mathrm{glob}}$ and $\boldsymbol{\sigma}_{\mathrm{glob}}$ denote the mean and standard deviation over all training atoms. Let $\mathcal{I}_k=\{i:g(a_i)=k\}$ denote the indices of training atoms belonging to family $k$. We standardize each posterior representation as
\begin{equation}
    \widetilde{\boldsymbol{\mu}}_i
    =(\boldsymbol{\mu}_i-\boldsymbol{\mu}_{\mathrm{glob}})
      \oslash\boldsymbol{\sigma}_{\mathrm{glob}},
    \label{eq:standardized_latents}
\end{equation}
where $\oslash$ denotes coordinate-wise division. For each observed family $k$, we compute its centroid in the standardized latent space and select the representation nearest to this centroid:
\begin{equation}
    \overline{\widetilde{\boldsymbol{\mu}}}_k
    =\frac{1}{|\mathcal{I}_k|}
      \sum_{i\in\mathcal{I}_k}\widetilde{\boldsymbol{\mu}}_i,
    \qquad
    i_k^*=\underset{i\in\mathcal{I}_k}{\arg\min}
      \left\|\widetilde{\boldsymbol{\mu}}_i
      -\overline{\widetilde{\boldsymbol{\mu}}}_k\right\|_2,
    \qquad
    \mathbf{m}_k=\boldsymbol{\mu}_{i_k^*}.
    \label{eq:medoid_selection}
\end{equation}
The term \emph{family medoid} in this paper refers to this centroid-nearest representation $\mathbf{m}_k$. 

\paragraph{Local scale estimation.}
For each family $k$, we select up to 1,000 same-family representations closest to its medoid, using Euclidean distance in the standardized latent space. We denote this neighborhood by $\mathcal{N}_k$ and use all available representations when the family contains fewer than 1,000 atoms.
We then estimate the local scale along each coordinate $j$ from the original, unstandardized latent representations:
\begin{equation}
s_{k,j}
=
\max\left\{
\operatorname{Std}_{i\in\mathcal{N}_k}(\mu_{i,j}),
10^{-5}
\right\}.
\label{eq:local_source_scale}
\end{equation}
The lower bound prevents the scale from vanishing. The resulting vector $\mathbf{s}_k$ captures the local variation along each latent dimension and determines the relative magnitude of the sampling perturbations. 

\paragraph{Bounded local sampling.}
Let $r_k$ denote the source component for family $k$ and $\rho_k$ its support radius. For distinct family medoids, set
\begin{equation}
    \rho_k
    =\beta\min_{\ell\in\mathcal{K}_{\mathcal{D}},\,\ell\ne k}
       \|\mathbf{m}_k-\mathbf{m}_{\ell}\|_2,
    \qquad \beta=0.2.
    \label{eq:source_radius}
\end{equation}
A sample from $r_k$ is generated as
\begin{equation}
    \begin{aligned}
        \boldsymbol{\epsilon}&\sim\mathcal{N}(\mathbf{0},\mathbf{I}),
        &\boldsymbol{\delta}_k&=\alpha\mathbf{s}_k\odot\boldsymbol{\epsilon},
        &\alpha&=0.1,\\
        \mathbf{z}^{(0)}&=\mathbf{m}_k+
        \boldsymbol{\delta}_k
        \min\!\left(1,
        \frac{\rho_k}{\|\boldsymbol{\delta}_k\|_2+10^{-12}}\right).
    \end{aligned}
    \label{eq:bounded_local_source}
\end{equation}
This operation projects a locally scaled Gaussian perturbation into a
bounded region centered at the family medoid. With $\beta=0.2<1/2$, the resulting source components have pairwise disjoint supports when the family medoids are distinct.

\subsection{Family-matched latent flow training}
\label{app:algorithms}
\label{app:hyperparameters}

For each training crystal $\mathcal{M}=(\mathbf{A},\mathbf{F},\mathbf{L})$,
we sample a target sequence $\mathbf{Z}^{(1)}\sim q_\phi(\mathbf{Z}\mid\mathcal{M})$ from the frozen VAE. Each target atom selects a source $\mathbf{z}^{(0)}_i\sim r_{g(a_i)}$. Consequently, the targets used to
train the flow are posterior \emph{samples}, whereas the family medoids and local scales in Appendix~\ref{app:source_construction} are estimated from posterior \emph{means}.

We draw $t\sim\mathcal{U}(0.01,0.99)$ and construct the linear path in Equation~\ref{eq:linear_path}. Consistently with Section~\ref{sec:preliminaries}, the network predicts the clean endpoint, and the velocity is obtained from that prediction:
\begin{equation}
    \widehat{\mathbf{Z}}^{(1)}=F_\theta(\mathbf{Z}^{(t)},t),
    \qquad
    \mathbf{v}_\theta(\mathbf{Z}^{(t)},t)
    =\frac{F_\theta(\mathbf{Z}^{(t)},t)-\mathbf{Z}^{(t)}}{1-t}.
    \label{eq:implemented_velocity}
\end{equation}
The sole training objective is
\begin{equation}
    \begin{aligned}
    \mathcal{L}_{\mathrm{FM}}
    &=\mathbb{E}\!\left[
      \frac{1}{Nd}\sum_{i=1}^{N}
      \left\|\mathbf{v}_\theta(\mathbf{Z}^{(t)},t)_i
      -(\mathbf{z}^{(1)}_i-\mathbf{z}^{(0)}_i)\right\|_2^2
      \right]\\
    &=\mathbb{E}\!\left[
      \frac{\|F_\theta(\mathbf{Z}^{(t)},t)-\mathbf{Z}^{(1)}\|_F^2}
           {Nd(1-t)^2}\right].
    \end{aligned}
    \label{eq:family_flow_loss}
\end{equation}
The implementation averages over unpadded atoms and all $d=8$ latent dimensions, followed by an average over crystals in the minibatch. 

Following \citet{joshi2025allatom}, we use a DiT-based flow network with 12 layers, a hidden dimension of 768, and 12 attention heads. An exponential moving average (EMA) of the flow parameters is updated after each optimization step and used for generation. The VAE remains frozen throughout flow training. Complete training and sampling settings are
provided in Table~\ref{tab:lfm_hyperparameters}.

\begin{table}[!htbp]
    \caption{Latent-flow settings for both datasets. The frozen VAE
    and the family sources are specific to each dataset.}
    \label{tab:lfm_hyperparameters}
    \centering
    \small
    \setlength{\tabcolsep}{4pt}
    \begin{tabularx}{\linewidth}{@{}Xll@{}}
        \toprule
        Setting & MP-20 & MPTS-52 \\
        \midrule
        Source noise $\alpha$ / radius factor $\beta$ & 0.1 / 0.2 & 0.1 / 0.2 \\
        DiT layers / hidden width / attention heads & 12 / 768 / 12 & 12 / 768 / 12 \\
        Training-time interval & $[0.01,0.99]$ & $[0.01,0.99]$ \\
        Optimizer / learning rate & AdamW / $10^{-4}$ & AdamW / $10^{-4}$ \\
        Weight decay / learning-rate scheduler & 0 / none & 0 / none \\
        Batch size / epochs & 128 / 2,000 & 128 / 2,000 \\
        Gradient-norm clipping & 10 & 10 \\
        EMA decay / update interval & 0.9999 / every step & 0.9999 / every step \\
        \bottomrule
    \end{tabularx}
\end{table}

\subsection{FamilySpace controllable sampling}
\label{app:sampling}

A requested family template $\mathbf{f}=(f_1,\ldots,f_N)$ specifies the family
of each source token, not its exact element identity. We independently sample $\mathbf{z}^{(0)}_i\sim r_{f_i}$ and transport the sequence using the EMA flow. The reported implementation uses an Euler grid from $t_0=0.01$ to $t_L=1$ with 100 grid points, giving 99 Euler updates. At each update,
\begin{equation}
    \mathbf{Z}_{\ell+1}
    =\mathbf{Z}_{\ell}
     +(t_{\ell+1}-t_{\ell})
       \mathbf{v}_{\theta_{\mathrm{EMA}}}
       (\mathbf{Z}_{\ell},t_{\ell}),
    \qquad \ell=0,\ldots,L-1,
    \label{eq:euler_sampling}
\end{equation}
where $\mathbf{v}_{\theta}$ denotes the backbone's sampling-time velocity. The initial state $\mathbf{Z}_0$ is the source sample. After the final Euler update, the network makes one additional clean-endpoint prediction at $t=1$. The frozen VAE decoder maps this prediction to atomic species, fractional coordinates, and lattice parameters.

\subsection{Source distribution ablation experiments}
\label{app:Source_ablation}

We investigate how different source centers affect crystal generation on MP-20. In our default setting, we select the training atom whose posterior mean is closest to the family centroid in the standardized latent space. We then use its posterior mean in the original latent space as the source center (Eq.~\ref{eq:medoid_selection}).
We compare the default medoid setting with two alternatives: (i) three random representative settings, where we randomly select one training atom from each family, using different representatives across the three runs; and (ii) evenly spaced centers, where ten centers are placed uniformly from $-10$ to $10$ along the first latent dimension, with all other dimensions set to zero. For all settings, we retain the original local scales multiplied by $0.1$ and apply the same radial clipping rule, with the radius recalculated for each set of centers (Eqs.~\ref{eq:source_radius} and~\ref{eq:bounded_local_source}). We train each flow model for 2,000 epochs using the same setting.

As shown in Table~\ref{tab:mp20_source_ablation}, our medoid setting achieves the highest stability rate and SUN among all settings. The evenly spaced centers achieve the highest uniqueness and novelty scores, but have lower stability and metastability rates. Random representative 2 achieves the highest MS and MSUN, with results close to those of our medoid setting. Overall, our medoid setting provides balanced performance across different metrics.

\begin{table*}[!t]
    \caption{Effect of the source centers on MP-20 generation.
    The best results are bold and the second-best results are underlined.
    }
    \label{tab:mp20_source_ablation}
    \centering
    \small
    \setlength{\tabcolsep}{5.0pt}
    \renewcommand{\arraystretch}{1.08}
    \resizebox{\textwidth}{!}{%
    \begin{tabular}{lcccccccc}
        \toprule
        \multirow{2}{*}{Method}
        & \multirow{2}{*}{Valid (\%) $\uparrow$}
        & \multicolumn{3}{c}{Uniqueness / novelty (\%) $\uparrow$}
        & \multicolumn{2}{c}{Stability (\%) $\uparrow$}
        & \multicolumn{2}{c}{Metastability (\%) $\uparrow$} \\
        \cmidrule(lr){3-5}
        \cmidrule(lr){6-7}
        \cmidrule(lr){8-9}
        & & U & N & UN & S & SUN & MS & MSUN \\
        \midrule
        FamilySpace (our medoid)
        & 89.48 & \underline{97.14} &  69.63 & 69.53
        & \textbf{6.26} & \textbf{1.38} & \underline{42.99} & \underline{15.31} \\
        FamilySpace (random representative 1)
        & \underline{90.88} & 95.72 & 71.41 & 70.63
        & \underline{5.86} & 1.01 & 42.27 & 14.14 \\
        FamilySpace (random representative 2)
        & \textbf{92.11} & 95.83 & 71.02 & 70.43
        & 5.77 & \underline{1.13} & \textbf{43.70} & \textbf{15.32} \\
        FamilySpace (random representative 3)
        & 88.38 & 96.43 & \underline{73.13} & \underline{72.40}
        & 5.57 & 1.07 & 40.93 & 14.73 \\
        FamilySpace (evenly spaced centers on $[-10,10]$)
        & 90.11 & \textbf{97.62} & \textbf{75.43} & \textbf{75.00}
        & 4.95 & 0.94 & 38.34 & 14.55 \\
        \bottomrule
    \end{tabular}%
    }
\end{table*}

\FloatBarrier
\section{Evaluation Details}
\label{app:evaluation_details}

\subsection{Compositional and structural validity}
\label{app:validity}

Following prior work~\citep{miller2024flowmm,zeni2025generative},
we evaluate validity from both compositional and structural perspectives. A crystal is considered valid only if it satisfies both criteria.

Compositional validity is evaluated with SMACT using the reduced stoichiometry. A composition is considered valid if at least one oxidation-state assignment satisfies charge neutrality and the Pauling electronegativity rules. Single-element structures are accepted by default, while all-metal alloys are allowed without enforcing ionic charge neutrality.

Structural validity is assessed on the MatterSim-relaxed structures rather than on the structures before relaxation. A structure is considered structurally valid if its minimum pairwise atomic distance under periodic boundary conditions is at least $0.5~\text{\AA}$ and its unit-cell volume is at least $0.1~\text{\AA}^3$.

\subsection{Uniqueness and novelty}
\label{app:matching}

Uniqueness and novelty are evaluated on relaxed structures using pymatgen's \texttt{StructureMatcher}~\citep{ong2013python,zeni2025generative}.
The length, site, and angle tolerances are \texttt{ltol=0.2},
\texttt{stol=0.3}, and \texttt{angle\_tol=5} degrees. 
For MP-20, novelty is evaluated against the Alex--MP reference distributed with MatterGen as \path{reference_MP2020correction.gz}. For MPTS-52, novelty is evaluated against the 27,380 structures parsed from its training CSV.
Uniqueness is computed within the full relaxed samples.

\subsection{Relaxation and thermodynamic stability}
\label{app:relaxation}

Generated crystals are relaxed with \texttt{MatterSim-v1.0.0-1M}
\citep{yang2024mattersim}. Both atomic positions and cell parameters are
optimized with FIRE and an \texttt{ExpCellFilter}. The stopping threshold
is $0.05~\mathrm{eV}/\text{\AA}$, with a maximum of 3,000 steps.
Finite final energies are processed with
\texttt{MaterialsProject2020Compatibility} and compared with the MP2020
hull reference for both datasets.

Using the reported energy above the reference hull, we define
\begin{equation}
    \mathrm{S}=\{j:E_{\mathrm{hull},j}\leq 0~\mathrm{eV/atom}\},
    \qquad
    \mathrm{MS}=\{j:E_{\mathrm{hull},j}\leq 0.1~\mathrm{eV/atom}\}.
    \label{eq:appendix_stability_thresholds}
\end{equation}
S denotes predicted stable structures. MS also includes structures
within $0.1~\mathrm{eV/atom}$ of the hull. The combined success
sets are
\begin{equation}
    \mathrm{SUN}=\mathrm{S}\cap\mathrm{U}\cap\mathrm{N},
    \qquad
    \mathrm{MSUN}=\mathrm{MS}\cap\mathrm{U}\cap\mathrm{N}.
    \label{eq:appendix_joint_metrics}
\end{equation}

\subsection{Family-level control metrics}
\label{app:control_metrics}

\paragraph{Evaluation setup.}
For each generated crystal, we compare the requested family composition
with the composition decoded by the frozen VAE. Metrics are computed over
all generated samples before validity filtering. For sample $j$, let
\[
\mathbf{c}_j=(c_{j,1},\ldots,c_{j,K}),
\qquad
\widehat{\mathbf{c}}_j
=(\widehat{c}_{j,1},\ldots,\widehat{c}_{j,K})
\]
denote the requested and decoded family-count vectors, respectively,
where $c_{j,k}$ and $\widehat{c}_{j,k}$ are the requested and decoded
numbers of atoms from family $k$ in sample $j$. We further define the
corresponding family sets as
\[
\mathcal{F}_j=\{k:c_{j,k}>0\},
\qquad
\widehat{\mathcal{F}}_j=\{k:\widehat{c}_{j,k}>0\}.
\]
All metrics depend only on composition and are independent of atom ordering.

\paragraph{Family composition accuracy.}
Let \(M\) denote the number of generated samples used to compute the control metrics.
\emph{Family-set accuracy} measures the fraction of samples that contain
exactly the requested families:
\begin{equation}
    R_{\mathrm{set}}
    =
    \frac{1}{M}\sum_{j=1}^{M}
    \mathbb{I}
    [\mathcal{F}_j=\widehat{\mathcal{F}}_j].
    \label{eq:app_exact_family_set}
\end{equation}

\emph{Family-count accuracy} additionally requires the number of atoms
from every family to match:
\begin{equation}
    R_{\mathrm{count}}
    =
    \frac{1}{M}\sum_{j=1}^{M}
    \mathbb{I}
    [\mathbf{c}_j=\widehat{\mathbf{c}}_j].
    \label{eq:app_exact_family_count}
\end{equation}

For experiments that specify family ratios, we report
\emph{Family-ratio accuracy}:
\begin{equation}
    R_{\mathrm{ratio}}
    =
    \frac{1}{M}\sum_{j=1}^{M}
    \mathbb{I}\!\left[
    \mathcal{F}_j=\widehat{\mathcal{F}}_j
    \ \land\
    \operatorname{red}(\mathbf{c}_j)
    =
    \operatorname{red}(\widehat{\mathbf{c}}_j)
    \right],
    \label{eq:app_exact_family_ratio}
\end{equation}
where $\operatorname{red}(\cdot)$ divides the nonzero family counts by
their greatest common divisor. All three accuracies are reported as
percentages.

\paragraph{Target-family fraction.}
For single-family control, we additionally report the
\emph{Target-family fraction}. Let
\[
\widehat{p}_{j,k}
=
\frac{\widehat{c}_{j,k}}
{\sum_{\ell=1}^{K}\widehat{c}_{j,\ell}}
\]
denote the decoded fraction of family $k$ in sample $j$.
For a target family $k$, the Target-family fraction is defined as
\begin{equation}
    \overline{p}_{\mathrm{target}}^{(k)}
    =
    \frac{1}{M_k}
    \sum_{j=1}^{M_k}
    \widehat{p}_{j,k},
    \label{eq:app_target_fraction}
\end{equation}
where $M_k$ is the number of generated samples for target family $k$.
It measures the average fraction of decoded atoms belonging to the
requested family and is reported as a percentage.

\FloatBarrier
\section{Additional Experiments}
\label{app:targeted_exploration}


\subsection{Within-family compositional diversity}
\label{app:control_diversity}

Family templates specify elemental families but do not fix the exact elements within each family. Therefore, a given multi-family template should allow multiple elemental compositions rather than collapsing to a single element choice for each family. For example, samples generated
from the Alkali family include Li, Na, K, Rb, Cs, and H, rather than consistently selecting only Li.
To verify that sampling from a specified family can produce diverse elemental choices, we analyze the within-family compositional diversity of the multi-family ratio control experiments in Table~\ref{tab:family_control}. As shown in
Table~\ref{tab:control_composition_diversity}, each family template produces many distinct elemental compositions. These results show that FamilySpace controls family-level composition without fixing the exact elemental composition within each family combination. 

\begin{table}[!htbp]
    \caption{
    Within-family compositional diversity in the multi-family ratio control experiments in Table \ref{tab:family_control}, with 10,000 generated samples for each family combination.
    \emph{Reduced compositions} column reports the number of distinct reduced elemental formulas, while \emph{Element sets} column reports the number of distinct element combinations, irrespective of stoichiometric ratios.
    }
    \label{tab:control_composition_diversity}
    \centering
    \small
    \setlength{\tabcolsep}{7pt}
    \begin{tabular}{@{}lrr@{}}
        \toprule
        Family combination & Reduced compositions & Element sets \\
        \midrule
        TM--Ch             & 4,337 & 1,950 \\
        Alkali--TM--Ch     & 2,418 & 1,284 \\
        AE--TM--Ch         & 3,004 & 1,885 \\
        RE--TM--Ch         & 2,435 & 1,899 \\
        Alkali--TM--Pn--Ch & 1,013 & 692 \\
        \bottomrule
    \end{tabular}
\end{table}

\subsection{Stability-oriented exploration protocols}
\label{app:targeted_protocols}

In Section~\ref{sec:De_novo}, we use selected family templates for stability-oriented exploration. We first choose six chemically motivated templates whose family compositions are commonly found in reported materials \citep{sirdeshmukh2001alkali,kumta1994rare,zinkevich2007thermodynamics,mitchell2002rare,hargittai2000molecular}. As an initial exploration, we generate 10,000 structures using a mixture of these six templates. The samples are distributed approximately uniformly across the six family ratios: four templates receive 1,667 samples each and two receive 1,666 samples each.

The results are summarized in Table~\ref{tab:targeted_sampling_appendix}. Among the six templates, Alkali--Halogen 1:1 and RE--Halogen 1:3 achieve the two highest MSUN rates. We therefore select these two templates for further evaluation and independently generate 10,000 additional structures for each. The focused runs confirm that both templates produce a high proportion of metastable candidates.

\begin{table*}[!htbp]
    \caption{
    Stability-oriented exploration on MP-20.
    $N$ denotes the number of generated samples.
    }
    \label{tab:targeted_sampling_appendix}
    \centering
    \small
    \setlength{\tabcolsep}{5pt}
    \begin{tabular}{@{}lrrrrrr@{}}
        \toprule
        Family template
        & $N$
        & UN (\%)
        & S (\%)
        & MS (\%)
        & SUN (\%)
        & MSUN (\%) \\
        \midrule

        Training templates (reference)
        & 10,000
        & 69.53
        & 6.26
        & 42.99
        & 1.38
        & 15.31 \\

        \midrule
\multicolumn{7}{l}{\emph{Six-template mixture}} \\
         Alkali--Halogen 1:1
        & 1,667
        & 51.71
        & 14.70
        & 90.16
        & 1.74
        & 42.11 \\

         AE--Halogen 1:2
        & 1,667
        & 62.75
        & 9.90
        & 70.13
        & 0.42
        & 34.43 \\

         Alkali--Ch 2:1
        & 1,667
        & 46.37
        & 21.96
        & 86.62
        & 2.10
        & 34.43 \\

         AE--Ch 1:1
        & 1,667
        & 19.80
        & 22.50
        & 77.38
        & 0.12
        & 4.56 \\

         RE--Ch 2:3
        & 1,666
        & 78.26
        & 9.01
        & 36.65
        & 0.93
        & 16.17 \\

         RE--Halogen 1:3
        & 1,666
        & 77.99
        & 8.17
        & 70.77
        & 6.10
        & 54.43 \\

        \midrule
        \multicolumn{7}{l}{\emph{Independent focused runs}} \\

        Alkali--Halogen 1:1
        & 10,000
        & 45.25
        & 13.56
        & 90.23
        & 0.70
        & 36.12 \\

        RE--Halogen 1:3
        & 10,000
        & 63.00
        & 7.87
        & 70.36
        & 4.97
        & 41.50 \\

        \bottomrule
    \end{tabular}
\end{table*}

\subsection{Application-oriented exploration}
\label{app:application_screening}

For application-oriented exploration, we generate 10,000 structures for
each of the five family templates in
Table~\ref{tab:application_motifs}. All generated structures are
evaluated using the same relaxation, hull-energy calculation, and
reference matching procedures described in
Appendix~\ref{app:evaluation_details}. We retain candidates satisfying
$\mathrm{MSUN}=\mathrm{MS}\cap\mathrm{U}\cap\mathrm{N}$ and examine
whether their chemical compositions have been reported in the literature
for relevant applications.

\begin{table*}[!htbp]
    \caption{
    Application-oriented exploration on MP-20.
    We generate 10,000 structures for each family template.
    U, N, UN, S, SUN, MS, and MSUN are reported as percentages.
    TM denotes Transition metal, AE Alkaline earth, RE Rare earth, Ch Chalcogens, and Pn Pnictogens, following Appendix~\ref{app:families}.
    }
    \label{tab:application_motifs}
    \centering
    \small
    \setlength{\tabcolsep}{7pt}
    \begin{tabular}{@{}lrrrrrrr@{}}
        \toprule
        Family template
        & U (\%) & N (\%) & UN (\%) & S (\%) & SUN (\%) & MS (\%) & MSUN (\%) \\
        \midrule

        TM--Ch
        & 90.25 & 89.01 & 85.93
        & 2.34 & 1.07 & 20.86 & 9.79 \\

        Alkali--TM--Ch
        & 87.63 & 88.93 & 84.30
        & 1.12 & 0.18 & 21.17 & 10.35 \\

        AE--TM--Ch
        & 82.64 & 85.27 & 78.98
        & 1.54 & 0.08 & 23.94 & 7.31 \\

        RE--TM--Ch
        & 87.37 & 89.48 & 83.84
        & 1.34 & 0.36 & 21.64 & 9.97 \\

        Alkali--TM--Pn--Ch
        & 94.85 & 98.73 & 94.40
        & 0.05 & 0.04 & 13.74 & 10.54 \\

        \bottomrule
    \end{tabular}
\end{table*}

\begin{figure*}[!htbp]
    \centering
    \includegraphics[width=\textwidth]{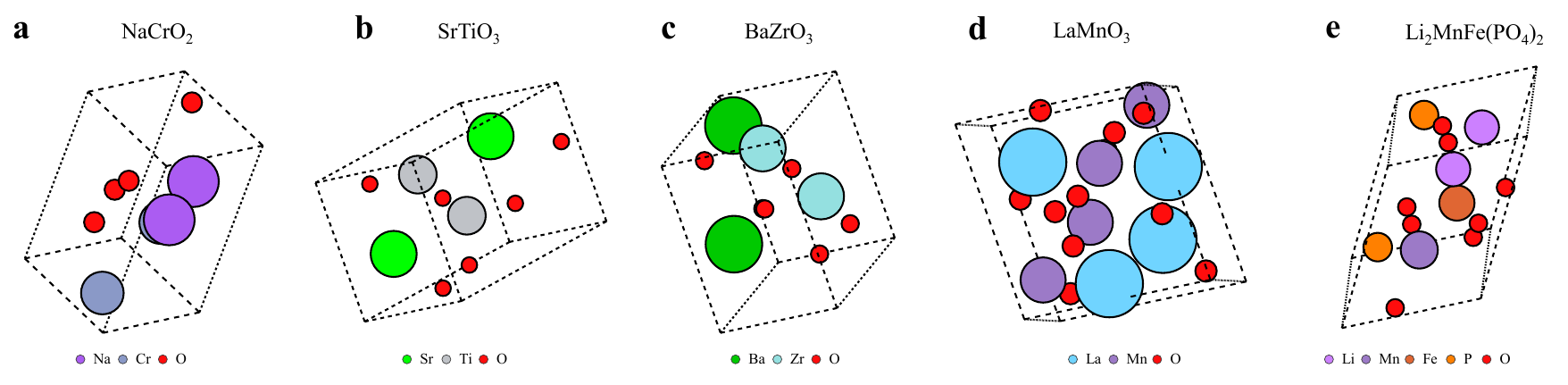}
    \caption{
    Representative generated MSUN candidates with
    application-relevant compositions.
    (a) $\mathrm{NaCrO_2}$ from the Alkali--TM--Ch template;
    (b) $\mathrm{SrTiO_3}$ and (c) $\mathrm{BaZrO_3}$ from the
    AE--TM--Ch template;
    (d) $\mathrm{LaMnO_3}$ from the RE--TM--Ch template; and
    (e) $\mathrm{Li_2MnFe(PO_4)_2}$ from the
    Alkali--TM--Pn--Ch template.
    }
    \label{fig:generated_structures}
\end{figure*}

Figure~\ref{fig:generated_structures} shows five representative
MSUN candidates whose compositions are associated with reported
functional materials.
The $\mathrm{NaCrO_2}$ composition in panel~(a) has been investigated
as a sodium-ion battery cathode, with nitrogen-doped carbon coating
used to improve its rate capability and cycling
stability~\citep{cai2025nitrogen}.
The compositions in panels~(b) and~(c), $\mathrm{SrTiO_3}$ and
$\mathrm{BaZrO_3}$, are associated with photocatalytic overall water
splitting~\citep{luo2025intrafacet} and proton-conducting ceramic
electrolyte hosts~\citep{tang2025phase}, respectively.
For $\mathrm{LaMnO_3}$ in panel~(d), recent work has investigated
unconventional cubic nanosheets for oxygen-evolution
electrocatalysis~\citep{dai2025microwave}.
The $\mathrm{Li_2MnFe(PO_4)_2}$ composition in panel~(e) is
stoichiometrically equivalent to
$\mathrm{LiMn_{0.5}Fe_{0.5}PO_4}$, which has been studied as a
lithium-ion battery cathode with engineered carbon
coatings~\citep{hou2025precursor}.